\documentclass[conference]{IEEEtran}
\IEEEoverridecommandlockouts
\usepackage{amsmath,amsfonts}
\usepackage{algorithmic}
\usepackage{algorithm}
\usepackage{array}
\usepackage[caption=false,font=normalsize,labelfont=sf,textfont=sf]{subfig}
\usepackage{textcomp}
\usepackage{stfloats}
\usepackage{url}
\usepackage{verbatim}
\usepackage{graphicx}
\usepackage{multirow}
\usepackage{cite}
\usepackage{xcolor}
\usepackage{cases}
\usepackage[scr=boondoxo,scrscaled=1.05]{mathalfa}
\usepackage[capitalise]{cleveref}

\begin{document}

\title{Enriching Neural Network Training Dataset to Improve Worst-Case Performance Guarantees\\

\thanks{This work is supported by the SYNERGIES project, funded by the European Commission Horizon Europe program, Grant Agreement No. 101069839, and by the ERC Starting Grant VeriPhIED, Grant Agreement No. 949899.}
}
\vspace{-0.5em}
\author{\IEEEauthorblockN{Rahul Nellikkath, Spyros Chatzivasileiadis}
\IEEEauthorblockA{\textit{Department of Wind and Energy Systems} \\
\textit{Technical University of Denmark (DTU)}\\
Kgs. Lyngby, Denmark \\
\{rnelli, spchatz\}@dtu.dk}
}
\vspace{-1.5em}



\maketitle

\begin{abstract}

Machine learning algorithms, especially Neural Networks (NNs), are a valuable tool used to approximate non-linear relationships, like the AC-Optimal Power Flow (AC-OPF), with considerable accuracy -- and achieving a speedup of several orders of magnitude when deployed for use. Often in power systems literature, the NNs are trained with a fixed dataset generated prior to the training process. In this paper, we show that adapting the NN training dataset \emph{during training} can improve the NN performance and substantially reduce its worst-case violations. This paper proposes an algorithm that identifies and enriches the training dataset with critical datapoints that reduce the worst-case violations and deliver a neural network with improved worst-case performance guarantees. We demonstrate the performance of our algorithm in four test power systems, ranging from 39-buses to 162-buses.

\end{abstract}

\begin{IEEEkeywords}
AC-OPF, Worst-Case Guarantees, Trustworthy Machine Learning, Explainable AI.
\end{IEEEkeywords}

\section{Introduction}
Machine learning algorithms, especially Neural Networks (NNs), are a valuable tool widely used to approximate non-linear relationships with considerable accuracy. An adequately sized NN with sufficient training data can be used to estimate complex non-nonlinear and non-convex power flow problems like the AC-Optimal Power Flow (AC-OPF) \cite{Convex} in a fraction of the time it would take to solve the exact problem. Considering AC-OPF is a valuable tool that the power system operators have to use numerous times to evaluate multiple uncertain scenarios and to prepare contingency plans for the daily safe operation of the power system, a well-trained NN would be a valuable tool that can save computational time and help the power system operators run orders of magnitude more scenarios. Moreover, these well-trained NNs could be used to either replace the existing convex approximations or relaxations of these non-convex non-linear problems \cite{ConvexRelax,VENZKE}, 
to warm start the actual problem \cite{warm}, or as a surrogate function in challenging optimization algorithms \cite{surrogate}. 

However, for the widespread adoption of the NN algorithms in power systems, it is essential to build an accurate approximation of the non-linear problems that do not violate the constraints of the power system.
In the case of approximating AC-OPF algorithms, a few researchers have shown that incorporating power flow constraint violations into the training process \cite{Sensitivity, Constraint1, Constraint2, Constraint3} can improve the accuracy of the predictions drastically. Previously, we had proposed a Physics-Informed Neural Network (PINN) \cite{PINN1}, which combined the KKT conditions of AC-OPF along with the power flow constraints to improve the performance of the NN \cite{MYAC}. Moreover, exciting research is happening on how the NN predictions could be altered to satisfy the power system constraints\cite{dc3}. These works have helped improve the accuracy and build trust in the NN predictions for AC-OPF predictions. 

Regardless, the performance of all these NN algorithms depends highly on the ability of the training data set to capture adverse points to the NN. Yet, a traditional NN training algorithm usually relies on a fixed training dataset, either randomly generated from the input domain or collected from a previously existing dataset, throughout the training process to generalize the problem. In a sizable system with multiple parameters to consider, one might require more than this traditional way of compiling training datasets to capture challenging adverse inputs to the NN that could lead to massive power system constraint violations once deployed. One of our previous works has shown the importance of enriching the training dataset with challenging input data points during training to improve its accuracy when predicting the N-1 small-signal stability margin of a wind farm \cite{closing}. 

Ultimately, to build trust in the NN's performance for a safety-critical application, such as power systems operation, some form of worst-case performance guarantees is required. Our previous work has shown how we can determine worst-case guarantees for AC \cite{MYAC} and DC-OPF \cite{Andreas} problems; and demonstrated how one could use the worst-case guarantees to select appropriate hyperparameters \cite{MYAC,MYDC} for the NN training which improve the worst-case performance. Still, as we show in this paper, we can substantially further improve the NN worst-case performance by enriching the training dataset.

In this paper, we illustrate how we can use the worst-case guarantees of the NN for AC-OPF to enrich the NN training data set with critical data points that improve the worst-case performance of the NN. Our contributions are as follows:
\begin{enumerate}
    \item We demonstrate how to use the worst-case guarantees of NN predictions, proposed in \cite{MYAC},  to enrich the NN training dataset during training to minimize the worst-case generation constraint violations of the NN.
    \item We demonstrate how we can use a simplified MILP problem to identify the region around the point that caused worst-case constraint violations to sample effectively multiple data points to the training dataset.
\end{enumerate}

This paper is structured as follows: Section II discusses AC-OPF and how we can use a power flow-informed NN to approximate the  AC-OPF solutions. Section III explains the proposed NN dataset enrichment algorithm.Section IV presents results from case studies, and Section V concludes. 

\section{Optimal Power Flow Algorithm and Power Flow Informed Neural Network}
This section describes the AC -OPF problem we used as a guiding application, presents the standard neural network architecture, and the power-flow-informed NN that will be used as a building block for the proposed NN training dataset enrichment algorithm.
\subsection{AC-Optimal Power Flow}
The objective function for reducing the cost of active power generation in a power system  with $N_g$ number of generators, $N_b$ number of buses, and $N_d$ number of loads can be formulated as follows:
\begin{equation}
    \min_{\mathbf{P}_g,\mathbf{Q}_g, \mathbf{v}} \quad \mathbf{c}^T_p\mathbf{P}_g
    \label{obj}
\end{equation}
where the vector $\mathbf{P}_g$ denotes the active power setpoints of the generators in the system, and $\mathbf{c}^T_p$ denotes the costs vector for the active power production at each generator. $\mathbf{Q}_g$ and $\mathbf{v}$ denote the reactive power setpoints of the generators and the complex bus voltages, respectively. The active and reactive power injection at each node $n \in N_b$, denoted by $p_n$ and $q_n$ respectively, can be calculated as follows:
\begin{align}
    p_n = p_n^g - &p_n^d  &\forall n\in N_b, \label{P_n} \\
    q_n = q_n^g - &q_n^d  &\forall n\in N_b, \label{Q_n}
\end{align}
where ${p}_n^g$ and ${q}_n^g$ are the active and reactive power generation, and ${p}_n^d$ and ${q}_n^d$ are the active reactive power demand at node $n$. The power flow equations in the network in cartesian cordinates  can be written as follows:
\begin{align}    
    p_n = \sum_{k=1}^{N_b} v_n^r(v_k^r G_{nk} - v_k^i B_{nk}) + v_n^i(v_k^i G_{nk} + v_k^r B_{nk}) \\
    q_n = \sum_{k=1}^{N_b} v_n^i(v_k^r G_{nk} - v_k^i B_{nk}) - v_n^r(v_k^i G_{nk} + v_k^r B_{nk})
\end{align}
where $v^r_n$ and $v^i_n$ denote the real and imaginary part of the voltage at node $n$. The conductance and susceptance of the line $nk$ connecting node $n$ and $k$ is denoted by $G_{nk}$ and $B_{nk}$ respectively. The power flow equations can be written in a more compact form by combining the real and imaginary parts of voltage into a vector of size $2N_b \times 1$ as $\mathbf v = [(\mathbf v^r)^T , (\mathbf v^i)^T]^T$ as follows \cite{Sensitivity}:
\begin{align}
    \mathbf{v}^T \mathbf{M}_{p}^n \mathbf{v} = &p_n        & \forall n\in N_b \label{PF_p}\\
    \mathbf{v}^T \mathbf{M}_{q}^n \mathbf{v} = &q_n            & \forall n\in N_b \label{PF_q}
\end{align}
where $\mathbf{M}_{p}^n$ and $\mathbf{M}_{q}^n$ are symmetric real valued matrices \cite{formulation}. 

The active and reactive power generation limits can be formulated as follows:
\begin{align}
    \underline{p}_n^g \leq & {p}_n^g \leq \overline{p}_n^g & \forall n\in N_g  \label{Pg_lim} \\
    \underline{q}_n^g \leq & {q}_n^g \leq \overline{q}_n^g & \forall n\in N_g \label{Qg_lim}
\end{align}

Similarly, the voltage and line current flow constraints for the power system can be represented as follows: 
\begin{align}
    \underline{\mathbf V}^n \leq \mathbf{v}^T \mathbf{M}_{v}^n \mathbf{v} &\leq \overline{\mathbf V}^n & \forall n\in N_b \label{V_lim} \\
    {\ell}_{mn} = \mathbf{v}^T \mathbf{M}_{i_{mn}} & \mathbf{v} \leq \overline{\mathbf \ell}_{mn} & \forall mn\in N_l \label{I_lim}
\end{align}
where $\mathbf{M}_{v}^n := e_ne^T_n + e_{N_b+n}e_{N_b+n}^T$ and $e_n$ is a $2N_b \times 1$ unit vector with zeros at all the locations except $n$. The squared magnitude of upper and lower voltage limit is denoted by $\overline{\mathbf V}^n$ and $\underline{\mathbf V}^n$ respectively. The squared magnitude of line current flow in line $mn$ is represented by ${\ell}_{mn}$ and matrix  $\mathbf{M}_{i}^{mn} = |y_{mn}|^2(e_m - e_n)(e_m -e_n)^T + |y_{mn}|^2(e_{N_b+m} - e_{N_b+n})(e_{N_b+m} - e_{N_b+n})^T$, where $y_{mn}$ is the line admittance of branch $mn$. Assuming the slack bus $N_{sb}$ acts as an angle reference for the voltage, we will have:
\begin{equation}
    v^i_{N_{sb}} = \mathbf{v}^T \mathbf{e}_{N_b + N_{sb}} \mathbf{e}^T_{N_b +N_{sb}} \mathbf{v} = 0 
    \label{V_sb}
\end{equation}

The constraints \eqref{P_n}-\eqref{Q_n},\eqref{PF_p}-\eqref{V_sb} and the objective function for the AC-OPF problem \eqref{obj} can be written in a more compact form as follows (for more details, see \cite{Sensitivity}): 
\begin{subequations}
\begin{align}
    \min_{\mathbf{v},\mathbf G} & \quad \mathbf{c}^T \mathbf G \label{pri_1}\\
    s.t \text{ } \mathbf{v}^T \mathbf{L}_l \mathbf{v} &= a_l^T \mathbf G +b_l^T \mathbf{D}, & l=1:L \label{pri_2}\\
    \mathbf{v}^T \mathbf{M}_m \mathbf{v} &\leq d_m^T\mathbf{D} + f_m, & m = 1 :M \label{pri_3}
\end{align}
\label{pri_full}
\end{subequations}
where $\mathbf{G} = [\mathbf{P}_g^T , \mathbf{Q}_g^T]^T$, $\mathbf{c}^T$ is the combined linear cost terms for the active power and, if necessary, reactive power generation. $\mathbf{D} = [\mathbf{P}_d^T , \mathbf{Q}_d^T]^T$ denotes the active and reactive power demand in the system. Following that, the equality constraints \eqref{PF_p}-\eqref{PF_q} and \eqref{V_sb} can be represented by the $\mathbf L=2N_b+1$ constraints in \eqref{pri_2}. Similarly, the inequality constraints \eqref{Pg_lim}-\eqref{I_lim} can be represented by the $\mathbf M=4N_g+2N_b+N_l$ constraints in \eqref{pri_3}.
\subsection{Neural Network for Optimal Power Flow Predictions}\label{SecPINN}
Neural Networks (NNs) are considered global approximators and can estimate the AC-OPF solution with significant accuracy when trained appropriately, and with sufficient training data. To achieve this, NN uses a group of interconnected hidden layers with multiple neurons to learn the relationship between the input and output layers. In the case of AC-OPF for generation cost minimization, the input layer will be the active and reactive power demand in the power system, and the output layer will be the optimal active and reactive power generation setpoints. 

Each neuron in a hidden layer will be connected with neurons in the neighboring layers through a set of edges. The information exiting one neuron goes through a linear transformation before reaching the neuron in the subsequent layer. Activation functions are used in every neuron to introduce non-linear relationships into the approximator. In this paper, we use the ReLU activation function in the hidden layers, as they have shown to accelarate convergence during training\cite{glorot}. 

\begin{figure}[htbp]
\centerline{\includegraphics[scale=.36]{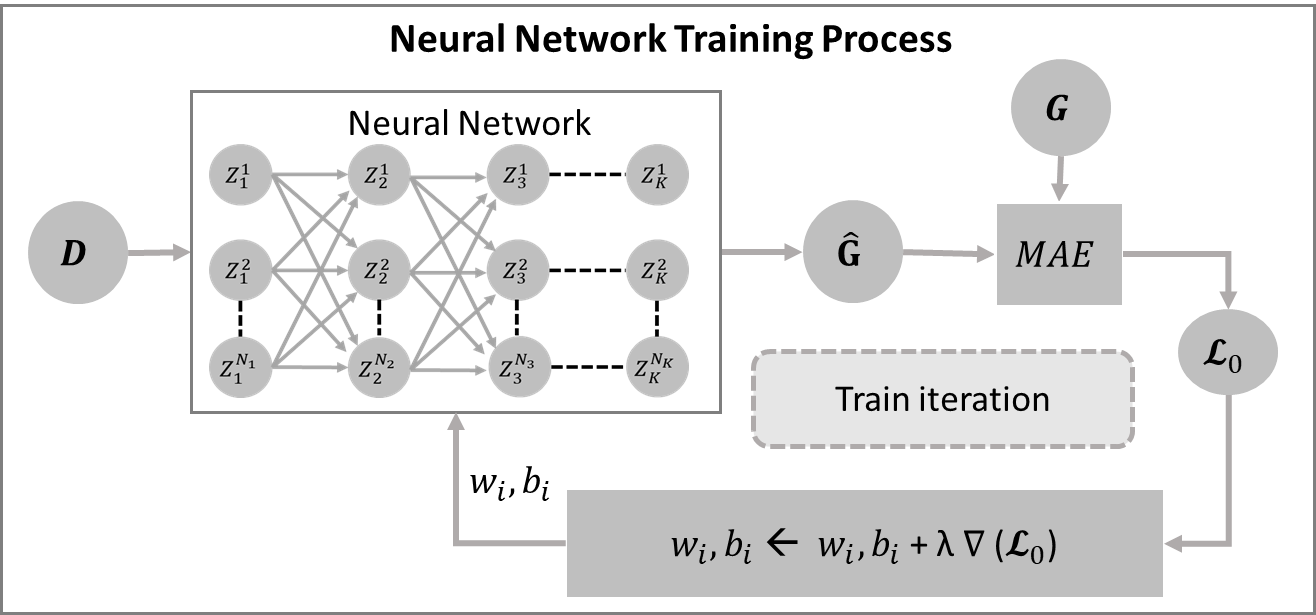}}
\caption{Illustration of the NN training architecture to predict the optimal active and reactive power generation setpoints ($\mathbf{\hat G}$) using the active and reactive power demand ($\mathbf{D}$) in the system as input: There are K hidden layers with $N_k$ neurons each. During training the weights $\mathbf{w_{k}}$ and biases $\mathbf{b_{k}}$ in the NN are optimized to minimize the average error ($\mathcal{L}_{0}$) in predicting the optimal generation setpoints ($\mathbf G$), calculated using \cref{L_0}.A learning rate of $\alpha$ controls the step size of the optimizer.}
\label{NN_basic}
\vspace{-1.2em}
\end{figure}

A standard NN, with $K$ number of hidden layers and $N_k$ number of neurons in hidden layer $k$, is shown in \cref{NN_basic}.

The information arriving at layer $k$ can be formulated as follows:
\begin{equation}
   \hat{\mathbf{Z}}_k = \mathbf{w_{k}}\mathbf{Z}_{k-1}+\mathbf{b_{k}}\label{NN1}
\end{equation}
where ${\mathbf{Z}}_{k-1}$ is the output of the neurons in layer $k-1$, $\hat{\mathbf{Z}}_k$ is the information received at layer $k$, $\mathbf{w_{k}}$ and $\mathbf{b_{k}}$ are the weights and biases connecting layer $k-1$ and $k$.

As mentioned, each neuron in the neural network uses the nonlinear ReLU activation function to accurately approximate the nonlinear relationships between the input and output layer. So, the output of each hidden layer in the NN can be represented as follows:
\begin{align}
    \mathbf{Z}_k &= \max( \hat{\mathbf{Z}}_k,0)\label{Relu}
\end{align}
The weights $\mathbf{w_{k}}$ and biases $\mathbf{b_{k}}$ in the NN are optimized to minimize the average error in predicting the optimal generation setpoints in the training data set, denoted by $\mathcal{L}_{0}$, which will be measured as follows:
\begin{equation}
    \mathcal{L}_{0} = \frac{1}{N} \sum_{i=1}^N | \mathbf{G}_i - \hat{\mathbf{G}}_i| \label{L_0}
\end{equation}
where $N$ is the number of data points in the training set, $\mathbf{G}_i$ is the generation setpoint determined by the OPF, and $\hat{\mathbf{G}}_i$ is the predicted NN generation setpoint. The back-propagation algorithm will be used to modify the weights and biases to minimize the average prediction error ($\mathcal{L}_{0}$), as shown in \cref{NN_basic}, during each iteration of the NN training. A learning rate of $\lambda$ controls the step size of the optimizer. 

\subsection{Power-Flow-Informed Neural Network}
A standard NN training procedure relies solely on the average error between its predictions and the data in the training dataset to achieve adequately accurate AC-OPF predictions. However, since these NN predictions should satisfy the power flow constraints given in \cref{pri_full}, we can use a power-flow-informed NN (PFNN) similar to the ones proposed in \cite{Sensitivity}, and our previous work \cite{MYAC} to improve the generalization capability of the NN. 

\begin{figure}[htbp]
\vspace{-0.5em}
\centerline{\includegraphics[scale=.3]{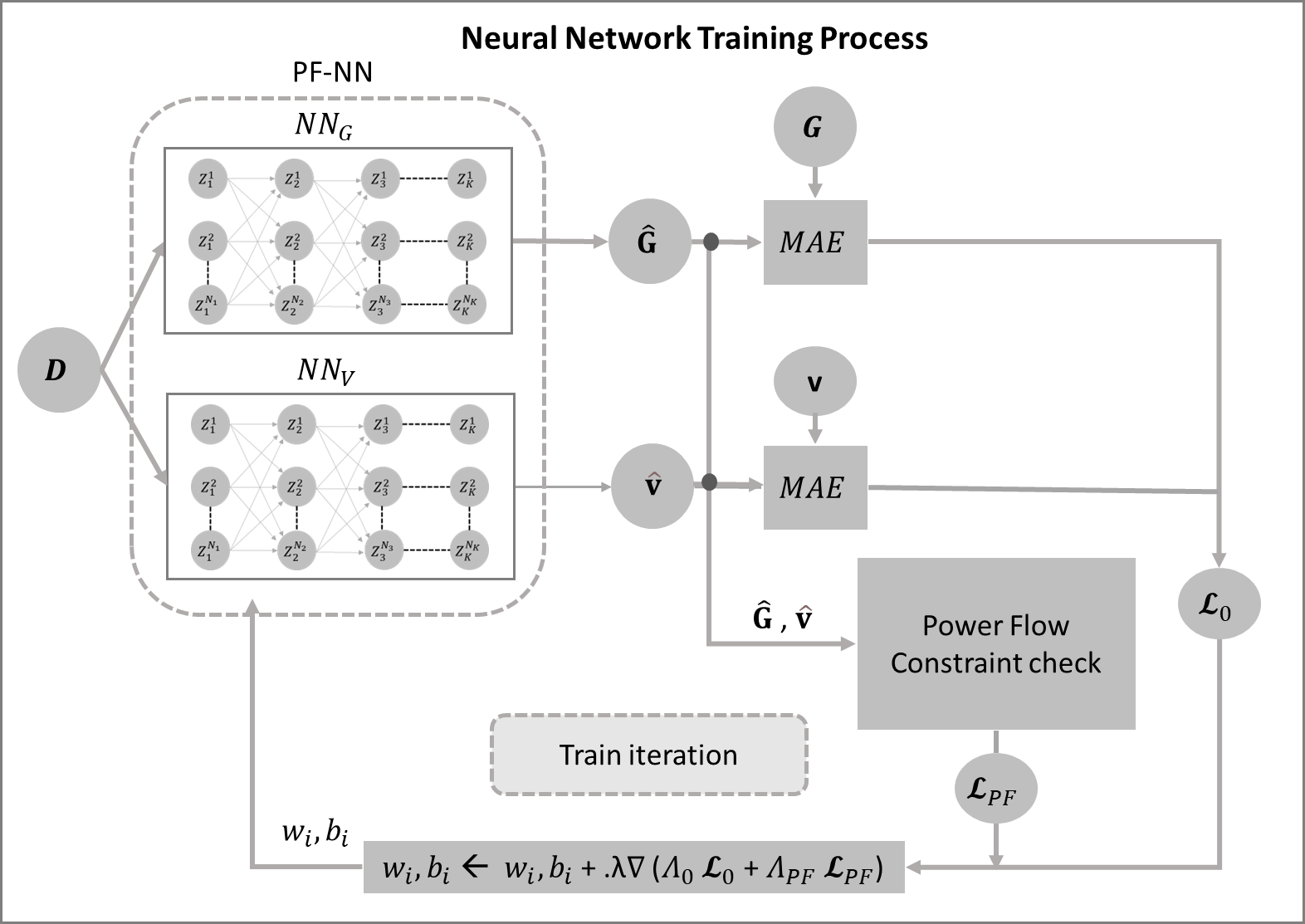}}
\caption{Illustration of the power-flow-informed NN (PFNN) training architecture to predict the optimal active and reactive power generation setpoints ($\mathbf{\hat G}$) using the active and reactive power demand ($\mathbf{D}$) in the system as input. There are two NNs, one for predicting $\mathbf{\hat G}$, denoted by $NN_G$, and the other for predicting the voltage in the system, denoted by $NN_v$ . During training, the weights $\mathbf{w_{k}}$ and biases $\mathbf{b_{k}}$ in the NN are optimized to minimize the average error ($\mathcal{L}_{0}$) in predicting the optimal generation ($\mathbf G$) and voltage setpoints ($\mathbf v$), and the power flow constraint violation($\mathcal{L}_{PF}$). See \eqref{L_pf} for the calculations.}
\label{PFNN}
\vspace{-0.5em}
\end{figure}

The structure of the power-flow-informed NN (PFNN) used in this work is given in \cref{PFNN}. Here an additional NN, denoted by $NN_v$, is used to approximate the real and imaginary part of the voltages in the system. Subsequently, both the predicted optimal generation setpoints and voltages in the system were analyzed to compute the power flow constraint violation, denoted by $\mathcal{L}_{PF}$, as follows:
\vspace{-0.1em}

\begin{subequations}
\small
    \begin{align} 
   &\mathcal{L}_{PF} = \frac{1}{N} \sum_{i=1}^N | \sigma_{eq} + \sigma_{ineq}| \\
\intertext{where,} \sigma_{eq} &= | \mathbf{v}^T \mathbf{L}_l \mathbf{v} - a_l^T \mathbf G - b_l^T \mathbf{D} |,\quad  &l=1:L \\
    \sigma_{ineq} &= \max( \mathbf{v}^T \mathbf{M}_m \mathbf{v} - d_m^T\mathbf{D} - f_m,0), &m = 1 :M
  \end{align} \label{L_pf}
\end{subequations}
\normalsize 
Thus the loss function for training the PFNN is formulated as follows:
\begin{equation}
   \mathcal{L}_{PFNN} =  \Lambda_0 \mathcal{L}_0 + \Lambda_{PF} \mathcal{L}_{PF}
\end{equation}
where $\Lambda_0$ and $\Lambda_{PF}$ are the weights given to the two loss functions $\mathcal{L}_{0}$ and $\mathcal{L}_{PF}$, respectively.

In our investigations, this configuration of NN with power flow constraint check has been shown to improve the average and worst-case performance of the NN without introducing a substantial computational burden. Still, the performance of the trained PFNN depends highly on the ability of the training dataset to capture adverse points to the  PFNN.
\section{Neural Network Data Enriching Using Worst-Case Constraint Violations} \label{EWCG}

The following section describes a NN dataset-enriching algorithm that uses worst-case generation constraint violations in regular intervals during training to identify the input regions causing the largest generation constraint violations and then enrich the training dataset with new training points from those regions. This shall help minimize the probability a NN prediction to lead to significant generation constraint violations. 
\subsection{Worst-Case Violations of the Neural Networks to Identify Adverse Examples} 
As mentioned earlier, we determine the input data with which we enrich our NN training dataset by identifying which are the NN inputs that cause the worst violations. The optimization problem to identify worst-case generation constraint violation is presented in \eqref{WCeq} \cite{MYAC}: 
\begin{subequations}
\vspace{-0.5em}
\begin{gather}
    \underset{\mathbf D}{\mathrm{max}} \quad v_g \label{WCeq1} \\
    v_g = \underset{ \mathbf D}{\mathrm{max}}(\mathbf{\hat{G} - \overline{G}}, \mathbf{\underline{G} - \hat{G}},0) \label{WCeq2}\\
    \text{s.t.} \text{ } \eqref{NN1}, \eqref{Relu}
\end{gather} \label{WCeq}
\end{subequations}
where $\mathbf{\overline{G}}$ and $\mathbf{\underline{G}}$ are the maximum and minimum active and reactive power generation bounds. However, the formulation for the ReLU activation function in \eqref{Relu} is nonlinear. So, the ReLU activation function is reformulated into a set of mixed-integer linear inequality constraints, as follows, to simplify the optimization problem \cite{Andreas}: 
\begin{subnumcases}
{\mathbf{Z}_k = \max(\hat{\mathbf{Z}}_k,0)\Rightarrow \label{MILP}}
\mathbf{Z}_k  \leq \hat{\mathbf{Z}}_k - \underline{\mathbf{Z}}_k  (1-\mathbf{y}_k) \label{RelU1} \\ 
\mathbf{Z}_k  \geq \hat{\mathbf{Z}}_k \label{RelU2}   \\
\mathbf{Z}_k  \leq \overline{\mathbf{Z}}_k \mathbf{y}_k  \label{RelU3}  \\
\mathbf{Z}_k   \geq \mathbf{0}  \label{Relu4}  \\
\mathbf{y}_k \in \{0,1\}^{N_k} \label{Relu5}.
\end{subnumcases}
where $\mathbf Z_k$ and $\hat{\mathbf{Z}}_k$ are the outputs and inputs of the ReLU activation function, $y^i_k$ is a binary variable, and $\underline{\mathbf{Z}}_k$ and $\overline{\mathbf{Z}}_k$ are the upper and lower limits of the input to ReLU. These limits should be sufficiently large to ensure they are not binding and small enough so the constraints are not unbounded. We used interval arithmetic to ensure a tighter bound (ref \cite{tjengg} and \cite{Andreas} for more information). If $\hat{\mathbf{Z}}_k$ is less than zero, then $y^i_k$ will be zero because of \eqref{RelU3}, and $Z^i_k$  will be constrained to zero. Else, $y^i_k$ will be equal to one, and $Z^i_k$ will be equal to  $\hat{\mathbf{Z}}_k$ due to \eqref{RelU1} and \eqref{RelU2}.   

The MILP optimization problem for identifying the specific input combination that could lead to a large generation constraint violation  (will be denoted by $D_{WC}$ from now on)  can be formulated as follows:
\begin{subequations}
\begin{gather}
    \underset{\mathbf D}{\mathrm{max}} \quad v_g  \Rightarrow v_g^{max}\label{WCVeq1} \\
    v_g = \mathrm{max}(\mathbf{\hat{G} - \overline{G}}, \mathbf{\underline{G} - \hat{G}},0) \label{WCeq2}\\
    \text{s.t.} \text{ } \eqref{NN1}, \eqref{MILP}
\end{gather} 
\end{subequations}
where $v_g^{max}$ is the worst-case generation constraint violation obtained after solving the optimization. 

By solving the optimization problem mentioned above for all the generators in the system, we can identify the specific $D_{WC}$ that caused the largest constraint violations. However, simply adding these  $D_{WC}$ to the dataset might not be adequate; instead, to properly enrich the dataset, it is also essential to identify the size of a region around  $D_{WC}$  that could lead to equally large constraint violations. Thus, we can collect multiple points from this region to enrich the dataset. 

The optimization problem below is used to fit a hypercube to the space around  $D_{WC}$ that could cause significant constraint violations.
\begin{subequations}
\begin{gather}
    \underset{\mathbf D}{\mathrm{max}} \quad d \label{Enrich1} \\
    d = \left\lVert D_0-D_{WC} \right\rVert_\infty \label{Enrich2}\\
    \text{s.t.} \text{ } v_g \geq \alpha \cdot v_g^{max} \\
     \eqref{NN1}, \eqref{MILP}
\end{gather} \label{WCen}
\end{subequations}
Here, the distance $d$ defines a hypercube around $D_{WC}$ that causes constraint violations more than $\alpha \cdot v_g$. We can choose a smaller or larger value of $\alpha$ to get a wider or tighter region around $D_{WC}$. 

In the optimization problem \eqref{WCen}, for $\alpha$ close to 1, we can assume the hypercube around $D_{WC}$ will be small. Thus we can consider a significant positive or negative $\hat{\mathbf{Z}}$ value will not change signs for any of the input data points inside the hypercube. Therefore, we can set those ReLU statuses, denoted by $\mathbf y_k$ in \eqref{MILP}, to always be active or inactive in the hypercube. This reduces the number of binary variables in the optimization problem and thereby reduces the computational complexity of the optimization problem. A threshold of 10 \% with respect to the  $\overline{\mathbf{Z}}$ or $\underline{\mathbf{Z}}$ was used to decide whether to fix $\mathbf y_k$  or not. That is, if the $\hat{\mathbf{Z}}$ value for NN input $D_{WC}$ was more than $0.1 \cdot \overline{\mathbf{Z}}$ or less than of $0.1 \cdot \underline{\mathbf{Z}}$, then those binary variables are set to active or inactive respectively in the hypercube. 

After obtaining the size of the hypercube, a random gaussian sampling, centered at $D_{WC}$, is used to sample new training data points from the hypercube. Unlike the original training dataset, we did not compute the optimal generation setpoints (i.e., G) for these additional data points to avoid using extra computational power to solve the AC-OPF problem. Instead, we used the power flow constraint violations in $\mathcal{L}_{PF}$ to train the neural network.

The proposed NN dataset enrichment algorithm is given in \cref{algo}.
\begin{algorithm}[]
\caption{Proposed NN Dataset Enriching algorithm}\label{algo}
\begin{algorithmic}[1]
\STATE {\textsc{TRAIN}}$(w_i, b_i)$
\vspace{3pt}
\STATE \hspace{0.5cm} Initialize the NN weights ($w_i$) and biases ($b_i$)
\vspace{3pt}
\STATE \hspace{0.5cm} \parbox[t]{210pt}{ Set the learning rate ($\alpha$), total number of epochs ($T$), number of iterations before accessing the worst-case violations to enrich the dataset for the first time ($T_{int}$) and number of iterations before enriching the dataset again ($T_{enr}$)}
\vspace{3pt}
\STATE \hspace{0.5cm} initialize t=1
\vspace{3pt}
\STATE \hspace{0.5cm} Repeat
\vspace{3pt}
\STATE \hspace{1 cm} \parbox[t]{200pt}{Calculate average losses $\mathcal{L}_0$ and $\mathcal{L}_{PF}$ in the training dataset}
\vspace{3pt}
\STATE \hspace{1 cm} \parbox[t]{200pt}{Update NN weights and biases $w_i$ and $b_i$ using the back propagation algorithm as follows:\\ $w_i, b_i = w_i,b_i + \alpha \nabla ( \Lambda_0 \mathcal{L}_0 + \Lambda_{PF} \mathcal{L}_{PF})$}
\vspace{3pt}
\STATE \hspace{1 cm} t=t+1
\STATE \hspace{0.5cm} until $t = T_{int}$
\vspace{3pt}
\STATE \hspace{0.5cm} Repeat
\vspace{3pt}
\STATE \hspace{1 cm} initialize $t_{enr}=1$
\vspace{3pt}
\STATE \hspace{1 cm} \parbox[t]{200pt}{Solve MILP optimization given in \cref{WCeq} is used to get the worst-case generation constraint violation $v_g^{max}$ and $D_{WC}$}
\vspace{3pt}
\STATE \hspace{1 cm} \parbox[t]{200pt}{Solve MILP optimization given in \cref{WCen} is used to get the hyper-cube around $D_{WC}$ and sample new data points from those hyper-cubes. Add these new points to the training set}
\vspace{3pt}
\STATE \hspace{1 cm} Repeat
\vspace{3pt}
\STATE \hspace{1.5 cm} \parbox[t]{200pt}{Calculate average losses $\mathcal{L}_0$ and $\mathcal{L}_{PF}$ in the \\training dataset}
\vspace{3pt}
\STATE \hspace{1.5 cm} \parbox[t]{200pt}{Update NN weights and biases $w_i$ and $b_i$ \\ using the back propagation algorithm as:\\ $w_i, b_i = w_i,b_i + \alpha \nabla ( \Lambda_0 \mathcal{L}_0 + \Lambda_{PF} \mathcal{L}_{PF})$}
\vspace{3pt}
\STATE \hspace{1.5 cm} $t_{enr}=t_{enr}+1$
\vspace{3pt}
\STATE \hspace{1.5 cm} t=t+1
\vspace{3pt}
\STATE \hspace{1 cm} until $t_{enr} = T_{enr}$
\vspace{3pt}
\STATE \hspace{0.5cm} until $t = T$
\end{algorithmic}
\label{alg1}
\end{algorithm}

\section{RESULTS \& DISCUSSION}
This section compares the worst-case generation constraint violations and the average performance of an NN with the proposed training dataset enrichment algorithm, denoted by WC-PFNN, against the power-flow-informed NN (PFNN) which uses a fixed training dataset. The results are analyzed for AC-OPF on 4 test systems: the 39-bus, 57-bus, 118-bus, and 162-bus systems from PGLib-OPF network library v19.05 \cite{PGLib}. The test system characteristics are given in \cref{TC}. 

In all instances, the input region for the active and reactive power demand was assumed to be convex, and at each node, the active and reactive power demand was considered to be between 60\% to 100\% of their respective nominal loading. This is a reasonable assumption used to simplify the dataset generation process. To give an idea about the size of each system, the sum of the maximum loading over all nodes for each system is given in \cref{TC}. 

\begin{table}[h]
\vspace{-1.0em}
\centering
\caption{TEST CASE CHARACTERISTICS}
\begin{tabular}{lllllll}
\hline\hline
\multirow{2}{*}{Test Case} & \multirow{2}{*}{$N_{b}$} & \multirow{2}{*}{$N_d$} & \multirow{2}{*}{$N_g$} & \multirow{2}{*}{$N_l$} & \multicolumn{2}{l}{Max total load} \\ \cline{6-7} 
                           &                          &                        &                        &                        & MW             & MVA             \\ \hline \hline
case39     & 39  & 21  & 10  & 46    & 6254   & 6626                                                    \\ \hline
case57     & 57  & 42  & 4  & 80   & 1251  & 1375                                                  \\ \hline
case118    & 118 & 99  & 19 & 186   & 4242 & 4537                                                      \\ \hline
case162    & 162 & 113 & 12 & 284   & 7239  & 12005                                                    \\ \hline \hline
\end{tabular}
\label{TC}
\vspace{-2em}
\end{table}

While generating the training data points, each node's active and reactive power demand was assumed to be independent. Then, for all cases, 10'000 sets of random input demand values were picked from the input domain using Latin hypercube sampling \cite{hypercube}. From these, 50\% were allocated to the training dataset, 20\% were assigned to the validation set, and the remaining 30\% were assigned to the unseen test dataset. The AC-OPF solver in MATPOWER \cite{MATPOWER} was used to obtain the optimal generation setpoints and the voltages at each bus.

A NN with three hidden layers and 20 nodes in each layer is used to predict the AC-OPF solutions. The ML algorithms were implemented using PyTorch \cite{Pytorch} and the Adam optimizer \cite{Adam}; a learning rate of 0.001 was used for training. WandB \cite{wandb} was used for monitoring and tuning the hyperparameters. The MILP problem for obtaining the worst-case violations and identifying the region was programmed in Pyomo \cite{pyomo} and solved using the Gurobi solver \cite{gurobi}. The NNs were trained in a High-Performance Computing (HPC) server with an Intel Xeon E5-2650v4 processor and 256 GB RAM. The code and datasets to reproduce the results are available online \cite{Code_git}.

\subsection{Comparing the Average and the Worst-Case Performance}
Both the PFNN and WC-PFNN algorithms, in all cases, achieved convergence in 600 iterations. After every 200 iterations during training, we evaluated the worst-case constraint violations, and in the case of WC-PFNN, a new 1000 data points were added to the training dataset. To compensate, 2000 random data points were added to the PFNN training dataset before starting the training process to have a comparable training dataset. 

For WC-PFNN and PFNN, we selected the weight $ \Lambda_{0}$ and $ \Lambda_{PF}$ for the loss functions $L_{0}$ and $L_{PF}$ that offered the lowest worst-case generation constraint violation using WandB. 

\begin{figure}[htbp]
\vspace{-0.8em}
\centerline{\includegraphics[scale=.45]{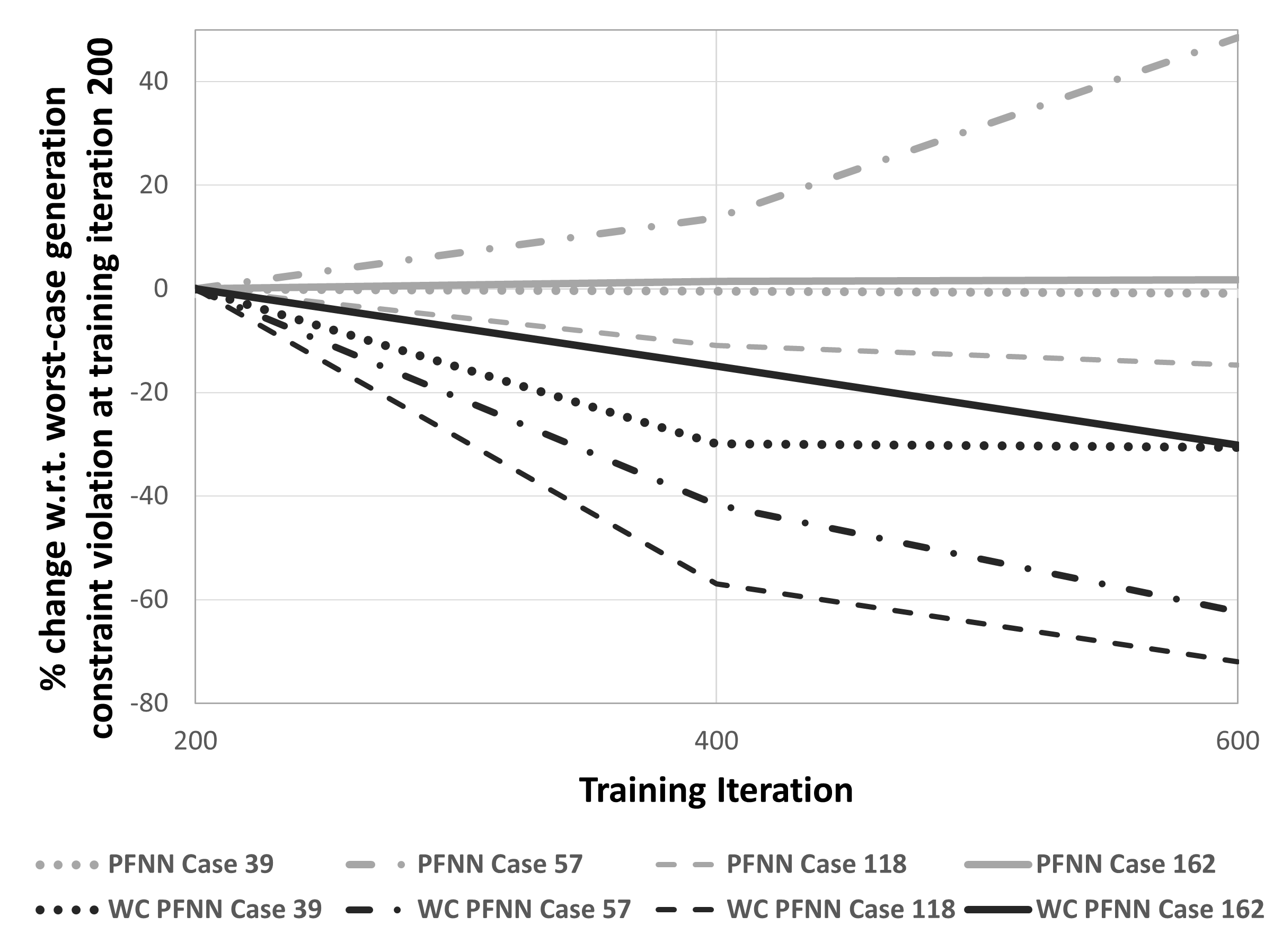}}
\vspace{-0.5em}
\caption{Change in worst case generation constraint violation with respect to the worst-case generation constraint violation at training iteration 200}
\label{res_fig}
\vspace{-0.5em}
\end{figure}

While comparing the worst-case performance over the training process, we can see in \cref{res_fig} that in all cases, the proposed NN dataset enriching algorithms (WC-PFNN) result in a steep reduction in worst-case guarantees. At the same time, the PFNN was only able to achieve a slight decline or, in case162 and case 57, even an increase in worst-case generation constraint violation during training. This highlights the importance of monitoring and enriching the NN dataset using worst-case constraint violations.

The mean absolute error (MAE) in an unseen test dataset, the worst-case generation constraint violation ($v_g$), and the width of the hypercube around $D_{WC}$, denoted by $d$ (as a fraction of the nominal loading at each node), caused by PFNN and WC-PFNN after training are given in \cref{Tab_res}. 
\begin{table}[]
\centering
\caption{Average and Worst-case performance}
\begin{tabular}{lllll}
\hline \hline
\multicolumn{2}{l}{Test Case}        & \begin{tabular}[c]{@{}l@{}}MAE \\      (\%)\end{tabular} & \begin{tabular}[c]{@{}l@{}}$v_g$\\      (MVA)\end{tabular} & $d$ \\ \hline \hline
\multirow{2}{*}{Case 39}  & PFNN    & 0.56                                                     & 6.08                                                       & 0.03             \\ \cline{2-5} 
                          & WC-PFNN & 0.52                                                     & 4.18                                                       & 0.02             \\ \hline
\multirow{2}{*}{Case 57}  & PFNN    & 1.2                                                      & 1280                                                       & 0.05             \\ \cline{2-5} 
                          & WC-PFNN & 1.1                                                      & 304                                                        & 0.01             \\ \hline
\multirow{2}{*}{Case 118} & PFNN    & 0.36                                                     & 1879                                                       & 0.04             \\ \cline{2-5} 
                          & WC-PFNN & 0.35                                                     & 600                                                        & 0.02             \\ \hline
\multirow{2}{*}{Case 162} & PFNN    & 1.4                                                      & 7552                                                       & 0.03             \\ \cline{2-5} 
                          & WC-PFNN & 1.2                                                      & 5212                                                       & 0.02             \\ \hline \hline
\end{tabular}
\label{Tab_res}
\vspace{-1.5em}
\end{table}

From \cref{Tab_res}, we see that WC-PFNN can reduce worst-case generation constraint violations by more than 30\% in all test systems. As a matter of fact, in case57, it achieves an up to 80\% reduction compared to PFNN with a fixed dataset. Moreover, it was observed that the hypercube around $D_{WC}$, which collects points that could cause significant constraint violations (see \eqref{WCen}), was also shrinking as we were including new points in the training set. So, the proposed method also made the NN safer in more regions in the input domain.
\section{Conclusion}
The objective of this paper is to demonstrate the importance of enriching the NN training dataset with critical datapoints in order to improve the NN worst-case performance. To determine these datapoints we use algorithms that accurately quantify the worst-case violations of the NN. This helps significantly improve the NN performance and enhances the trust in NN for safety-critical applications. In this paper, we show that by enriching the NN training dataset using the worst-case generation constraint violations of an AC-OPF problem during training, we can improve the worst-case performance of the NN drastically. We test the proposed algorithm on four different power systems, ranging from 39-buses to 162-buses and we show that we can achieve up to 80\% reduction in worst-case performance. In future work, we plan to integrate this approach with our recent work on designing a neural network training procedure that achieves best average performance and minimizes the worst case violations at the same time \cite{MIN-WC}.
\bibliographystyle{IEEEtran}
\bibliography{references}

\vfill

\end{document}